\documentclass[letterpaper, 10 pt, conference]{ieeeconf}  

\IEEEoverridecommandlockouts
\usepackage{cite}
\usepackage{amsmath,amssymb,amsfonts}
\usepackage{algorithmic}
\usepackage{graphicx}
\usepackage{textcomp}
\usepackage{booktabs}
\usepackage{xcolor}
\usepackage{svg}
\usepackage{subfig}

\title{\LARGE \bf Toward Understanding Key Estimation in Learning Robust Humanoid Locomotion
}

\author{Zhicheng Wang$^{1}$ Wandi Wei$^{1}$ Ruiqi Yu$^{1}$ Jun Wu$^{1,2}$ Qiuguo Zhu$^{*1,2}$%
\thanks{$^{1}$ The authors are with Institute of Cyber-Systems and Control, Zhejiang University, 310027, China}%
\thanks{$^{2}$ Qiuguo Zhu and Jun Wu are with State Key Laboratory of Industrial Control Technology, 310027, China}%
\thanks{$^*$ Qiuguo Zhu ({\tt\small qgzhu@zju.edu.cn}) is the corresponding author.}%
\thanks{$^\dagger$ This work was supported by the National Key R\&D Program of China (Grant No. 2022YFB4701502), the ”Leading Goose” R\&D Program of Zhejiang(Grant No. 2023C01177), the Key R\&D Project on Agriculture and Social Development in Hangzhou City (Asian Games) (Grant No. 20230701A05), and the Key Research Project of Zhejiang Lab (Grant No. 2021NB0AL03)}%
}

\begin{document}
\maketitle
\thispagestyle{empty}
\pagestyle{empty}

\begin{abstract}
Accurate state estimation plays a critical role in ensuring the robust control of humanoid robots, particularly in the context of learning-based control policies for legged robots. However, there is a notable gap in analytical research concerning estimations. Therefore, we endeavor to further understand how various types of estimations influence the decision-making processes of policies. In this paper, we provide quantitative insight into the effectiveness of learned state estimations, employing saliency analysis to identify key estimation variables and optimize their combination for humanoid locomotion tasks. Evaluations assessing tracking precision and robustness are conducted on comparative groups of policies with varying estimation combinations in both simulated and real-world environments. Results validated that the proposed policy is capable of crossing the sim-to-real gap and demonstrating superior performance relative to alternative policy configurations.

\end{abstract}


\section{INTRODUCTION}
Humanoid robots hold immense potential and practical value as general-purpose robots. However, their high degrees of freedom and system complexity pose significant challenges to achieving stable control.

Over the past few decades, researchers have proposed various methods to enhance the mobility of robots. Classical control methods can achieve stable static motion\cite{raibert1986legged, Westervelt2003hzd, collins2005efficient}, and furthermore, optimization-based strategies can generate dynamic behaviors while adhering to constraints\cite{Jacob2019mc, Huang2023mc, Yukai2019mc, Gibson2022cassieMc}. The most notable exemplar is the Boston Dynamics Atlas, capable of smoothly performing parkour, back-flipping, and object manipulation\cite{altas2023mc}.

With the advancement of computational power, learning-based methods have become popular in legged robot control and have also achieved remarkable achievements. These learning methods were initially validated on quadruped robots, exhibiting the capability to traverse diverse terrains\cite{ANYmalSciRo1, ANYmalSciRo2, ANYmalSciRo3,margolisyang2022rapid} and are beginning to incorporate visual information\cite{agarwal2022Egocentric,loquercio2022CMS}. On bipedal robots, learning policies are also capable of executing fundamental motions\cite{CassieVonMises, Siekmann2021Cassie, Duan2022SteppingStones, Ilija2023HumanoidTransformer}. Additionally, there are endeavors that combine models with learning methods\cite{clavera2018model, Hybrid, batke2022optimizing}.

In theory, the more information a policy acquires, the better its performance, yet not all data is readily available in real-world robot deployment, such as contact forces, deformable surfaces, and precise velocity. These data are referred to as privileged information. Recently, many new learning-based methods have introduced additional modules to extract privileged information from available observational data to implicitly estimate the robot's states and surrounding environment. A classic framework is RMA\cite{kumar2021rma}. Other works are using explicit estimating methods\cite{estimatorNet}. However, the information extracted by RMA is in the form of latent vectors, which are difficult to interpret, and the two-stage training process is inefficient. The new framework introduces explicit estimates based on latent vectors, such as the robot's linear velocity\cite{nahrendra2023dreamwaq} and height map\cite{cheng2023parkour}, further improving the robot's motion performance and terrain adaptability.


\begin{figure}
    \centering
    \includegraphics[width=8.0 cm]{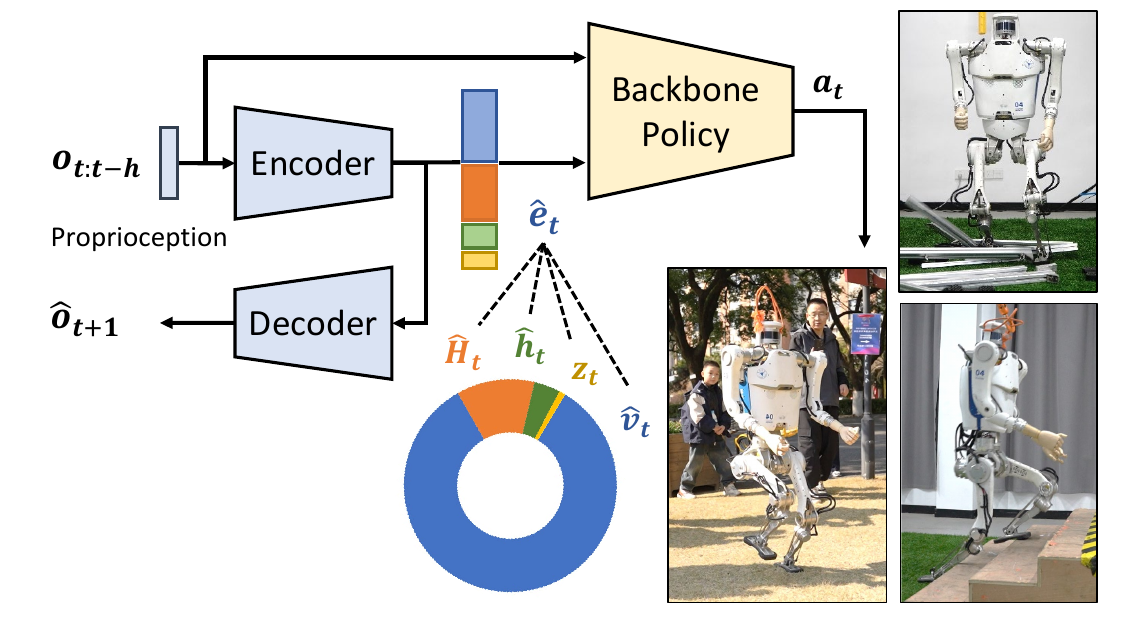}
    \caption{Overview of key estimation policy. By quantifying the importance of the explicit estimation states and designing the key estimation architecture, the policy achieves real-world blind locomotion with a real Wukong-IV humanoid.}
    \label{Fig_Showcase}
\end{figure}

\begin{figure*}[!ht]
    \centering
    \includegraphics[width=16cm]{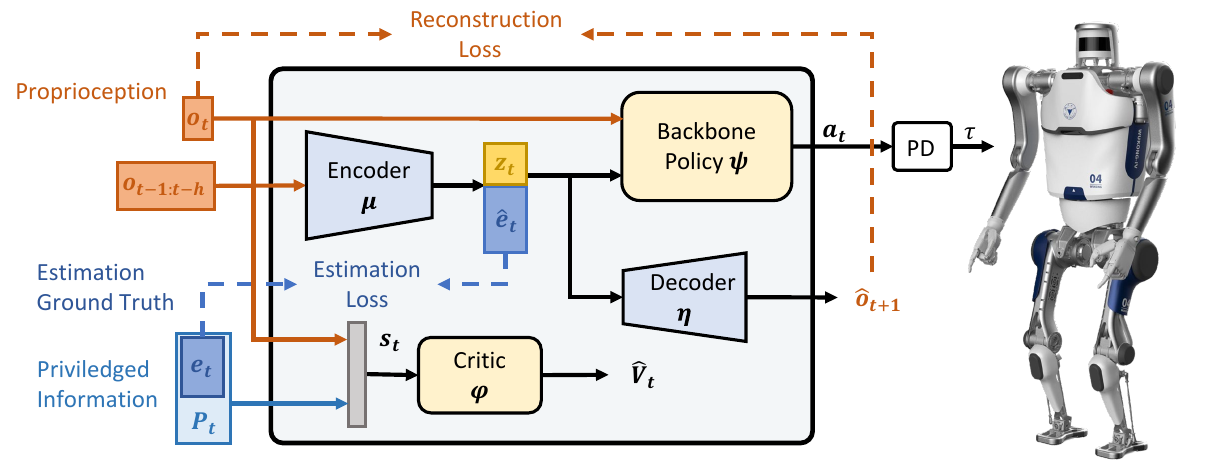}
    \caption{Architecture of the proposed policy. The actor consists of an auto-encode $(\mu, \eta)$ and a backbone policy $\psi$. The encoder $\mu$ takes in 0.5s-long historical proprioception and generates estimations, implicit encoding ${z_t}$ and explicit encoding ${\hat{e}_t}$. The decoder $\eta$ reconstruct the current proprioception using ${z_t}$ and ${\hat{e}_t}$. The ${\hat{e}_t}$ is fitted to the true values of corresponding physical variables ${e_t}$. Both encodings, along with current proprioception ${o_t}$, serve as input to the backbone policy, resulting in actions. The critic's input includes ${o_t}$ and privileged information ${P_t}$ that includes ${e_t}$ and other useful data.}
    \label{Fig_Architecture}
\end{figure*}

For humanoid robots, it is speculated that diverse estimation methods could enhance policy-driven motion performance. In the interdisciplinary domain of robotics and AI, the meticulous design and selection of estimation modules are vital. These modules not only define the breadth and depth of information assimilated by the learning algorithm but also leverage the extensive prior knowledge accumulated within the field of robotics.
Consequently, several key questions arise: "Should estimations be explicit or implicit?" "Which variables should be estimated?" and "How do estimations assist the policy to finish the locomotion task?" While some research has investigated the significance of proprioception\cite{yang2020multi}, quantitative analyses of the importance of estimation remain largely unexplored.

In this paper, we aim to explore the effect of estimation further. First, we trained a policy with the estimation of various privileged information. Through saliency analysis, we ranked the importance of each estimation and proposed an optimal combination. Then, we conducted various tests on policies with different estimations. The results indicate that the policy with the optimal combination of estimations achieves the best overall performance. During real robot deployment, our policy can walk through challenging environments like stairs, slopes, and obstacles when tracking velocity commands.

The main contributions of this work are:
\begin{itemize}
    \item Quantitative analysis of the influence of the estimation variables on the performance of learned policies, and proposed the optimal combination.
    \item A controllable and adaptive framework for learning humanoid locomotion with the proposed effective estimation scheme based on asymmetric actor-critic.
    \item The proposed learning framework and estimation methodology are tested in the real world and prove to be capable of adapting to outdoor environments.
\end{itemize}

The structure of this paper is as follows: Section \ref{Methods} describes the method to train our policy. Section \ref{experiment setup} presents the training process and the experiments to evaluate our policy, including results and analysis. Section \ref{Conclusion} summarizes the paper and states the future work.

\section{METHODOLOGY}
\label{Methods}

\subsection{Asymmetric Policy Architecture}

To avoid multiple training stages and imitation inaccuracies in teacher-student training architecture, this work follows an asymmetric actor-critic structure. The actor policy only has access to a 0.5s-long realistic observation history that contains delayed noisy proprioceptive information and commands. The critic policy has access to all kinds of states. The actor is composed of an auto-encoder $(\mu, \eta)$ that predictively reconstructs the proprioceptive information, and a backbone policy $\psi$ that conditions joint-level position targets on the latest observation and estimation encodings. All components are fully connected neural networks. The size of hidden layers is described in Appendix \ref{Appendix_nn}.

To train every part of the framework, the training loss function is formulated as a weighted sum of policy gradient loss, observation prediction loss, and estimation loss. In this work, the policy gradient loss is computed with Proximal Policy Optimization(PPO)\cite{PPO}.

The training process and the architecture of the policies are shown in Fig. \ref{Fig_Architecture}.

\subsection{State and Action}
States are categorized into three types: observation, privilege, and command. The observation, denoted as {$O_t\in\mathbb{R}^{42}$} includes the proprioceptive accessible from the actual robot. The observation is composed of the gravity vector, body angular velocity, joint position, joint velocity, and the previous action. The privileged information, represented as {$P_t\in\mathbb{R}^{103}$}, pertains to data that is intricate to acquire in the real world, including ground truth of body linear velocity, body height, small heightmap around the robot's feet and large heightmap around the robot's base. Except for a larger heightmap around the robot, all physical quantities in the privileged information can be explicitly estimated. User command $\in\mathbb{R}^{7}$ includes gait signal, desired linear velocity, and desired yaw angular velocity. All commands are expressed in the robot frame.

Action {$a_t$} is the desired joint position, which is executed by a subsequent PD controller. The policy runs and generates action at 100 Hz. The PD controller runs at 1kHz.

\subsection{Reward}
Bell-shape kernel functions are introduced into reward design to encourage the policy to survive. The adopted kernel functions can be formulated as:

\begin{eqnarray}
G_{\alpha,\sigma}(x)&=&\alpha \exp{(\frac{x}{\sigma})^2} \\
C_{\alpha,\beta,\sigma}(x)&=&\alpha((\frac{x}{\sigma})^{2\beta}+1)^{-1}, \beta=1,2,...
\end{eqnarray}
where $G_{\alpha,\sigma}(x)$ represents the Gaussian kernel function, while $C_{\alpha,\beta,\sigma}(x)$ denotes the generalized Cauchy kernel function, with $\alpha$, $\beta$, and $\sigma$ serving as adjustable coefficients. The Gaussian function is frequently utilized in reward design due to its advantageous gradient behavior around zero, which enhances precision in tracking tasks. However, its gradients may become insignificantly small in distant regions. Conversely, the generalized Cauchy kernel function, characterized by its heavy-tailed feature, offers viable gradients for learning even when $x$ is far from zero. Notably, this function possesses a stationary point at $(\pm \sigma, 0.5)$, simplifying the adjustment of the scaling coefficient $\sigma$. Moreover, the additional order coefficient $\beta$ allows for modifications to the kernel's shape. A larger $\beta$ value tends to shape the function more closely to a rectangular function, thereby imposing lesser penalties when $x$ falls below $\sigma$.

The reward items in this work can be divided into the following categories: (1) \emph{base command tracking}, (2) \emph{gait}, (3) \emph{smoothness and energy saving}.

\subsubsection{Base command tracking}
The commands include base height, base orientation, base linear, and angular velocity. The reward functions are defined as:
\begin{eqnarray}
r_{v}&=&G_{0.1, 0.02}(||v_{t}^*-v_{t}||) \\
r_{\omega}&=&G_{0.1, 0.02}(||\omega_{t}^*-\omega_{t}||) \\
r_{r}&=&G_{0.1, 0.0025}(1-R_{2,2}^2)\\
r_{h}&=&G_{0.2, 0.02}(|h^*-h|)
\end{eqnarray}
where {$h$} is the vertical distance from the base to the ground, {$v\in\mathbb{R}^3$} represents the linear velocity, and {$R_{2,2}$} is the bottom right corner value in the base rotation matrix, {$1-R_{2,2}$} measures the deviation between the Z-axis of robot base and the world frame.

\subsubsection{Gait}
We inherit the gait-related command and reward design from \cite{gaitSig} to generate a controllable locomotion pattern. It penalizes contact force during the swing phase and foot movement during the contact phase. The reward functions are defined as:
\begin{eqnarray}
r_{eVel}&=&C_{0.1, 1, 8}(Q_{v,l}V_l+Q_{v,r}V_r)\\
r_{eFrc}&=&C_{0.1, 1, 8}(Q_{f,l}F_l+Q_{f,r}F_r)
\end{eqnarray}
where {$I_{v/f,l/r}$} represents the gait force/velocity penalization coefficient for each foot, {$V_{l/r}$} is the foot velocity of the corresponding foot, and {$F_{l/r}$} is the contact force of corresponding foot.

\subsubsection{Smoothness and energy saving}
We hope our policy can optimize the impact force on the foot, joint torque smoothness, joint velocity smoothness, and cost of transport(CoT). We adopt The reward functions are defined as:
\begin{eqnarray}
r_{i}&=&C_{0.1, 3, 0.2}(\frac{||F_t-F_{t-1}||}{mg})\\
r_{\tau}&=&C_{0.1, 2, 160}(||\tau_t-\tau_{t-1}||)\\
r_{\dot{q}}&=&C_{0.1, 1, 8}(\frac{||\dot{q}_{t}-\dot{q}_{t-1}||}{||v_{t}||})\\
r_{CoT}&=&C_{0.1, 3, 1.6}(\frac{\tau_t\cdot \dot{q}_{t}}{mg||v_{t}||})
\end{eqnarray}
where {$F_t$} represents the foot force at step {$t$}, {$mg$} is the gravity of robot, {$\tau_t$} is the joint torque at step {$t$}, {$\dot{q}_{t}$} is the joint velocity at step {$t$}. CoT is a fairer cross-platform efficiency evaluation indicator\cite{CoT}, with better results than single penalty joint torque.

\begin{figure*}[t]
    \vspace{1mm}
    \centering
    \includegraphics[width=15.5 cm]{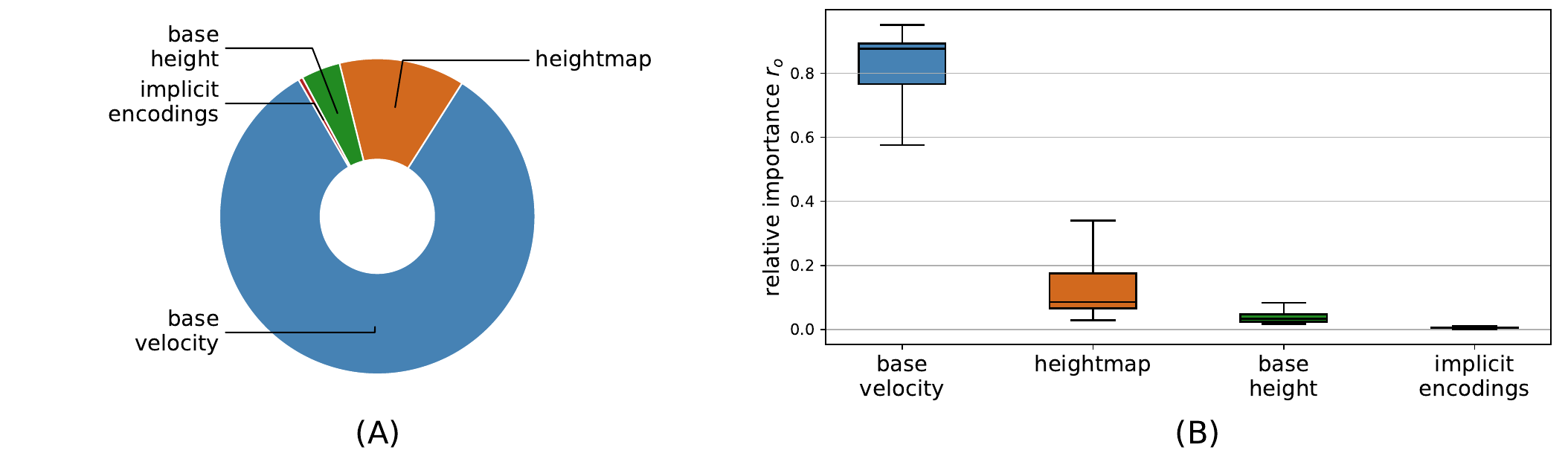}
    \caption{Saliency analysis of the estimation states. (A) Pie chart of the estimations' average relative importance. (B) Box plot of the ranges of the relative importance for all samples. The colored box refers to the range between 25\% and 75\% samples, the horizontal line is the median number, and the error bar shows the boundaries where $p<0.05$.}
    \label{Fig_saliency}
\end{figure*}

\subsubsection{Saliency Analysis}

To understand the importance of the estimation terms quantitatively, we adopted integrated gradients from explainable artificial intelligence as saliency analysis metrics. Given $N$ timesteps of input $x\in\mathbb{R}^n$ and a policy $F(x)\in\mathbb{R}^m$, the input $x$ can be categorized into $\chi$ groups according to the physics meaning, and the dimension of the category is denoted as $h$, the saliency metrics can be formulated as:

\begin{equation}
G_{x_{i,t}} = \sum_{j=1}^{m}{|\frac{x_{i,t}-\hat{x}_{i,t}}{p}\sum_{k=1}^{p}{\frac{\partial F_j(\frac{p-k}{p}\hat{x}_{t}+\frac{k}{p}x_t)}{\partial x_{it}}}}|
\end{equation}

\begin{eqnarray}
S_d(x_{i,t}) &=& \max{(G_{x_{i,t}}-\epsilon, 0)}\\
\epsilon &=& \frac{1}{nN}\sum_{i=1}^{n}\sum_{t=1}^{N}G_{x_{i,t}}\\
S(x_{i,t}) &=& \frac{S_d(x_{i,t})}{\max(S_d(x_{i,t}))} \\
I_i &=& \sum_{t=1}^{N}S(x_{i,t}) \\
I_o &=& \frac{1}{h}\sum_{q=1}^{h}I_q \\
\iota_o &=& \frac{I_o}{\sum_{k=1}^{\chi}I_k}
\end{eqnarray}
where $G_{x_{i,t}}$ represents integrated gradient, $S_d$ and $S$ represent the absolute and relative saliency, $I_i$ and $I_o$ are element-wise and total importance, and $\iota$ is the relative importance of the input category $o$. $\hat{x}_{i,t}$ is the nominal input which is usually zero, $p=25$ is the integral horizon.

\subsection{Training Environment Design}

We adopted the game-inspired terrain curriculum from \cite{isaac}. A sequence of complex terrain is generated from easy to hard. Once the robot moves far enough from the starting point in an episode, it will be spawned on the next-level terrain in the following episode. Conversely, the robot that failed to survive long enough will be spawned on an easier terrain in the next episode.

To achieve better real robot performance under a variety of commands, randomized commands, and domain randomization are applied to the training environment. Environment parameters and commands are sampled from associated uniform distributions at the beginning of every episode. Environment parameters include initial robot pose, mass payload, ground friction coefficient, motor strength, and observation latency. Commands include linear velocity, facing direction, and gait signal. Furthermore, a stochastic noise is added to the normalized proprioceptive observation to imitate sensor noise in real robots. The ranges of random parameters can be found in Appendix \ref{Appendix_rand}.

\section{experiment and result}
\label{experiment setup}

\subsection{Platform description}

\begin{figure}
    \centering
    \includegraphics[width=8 cm]{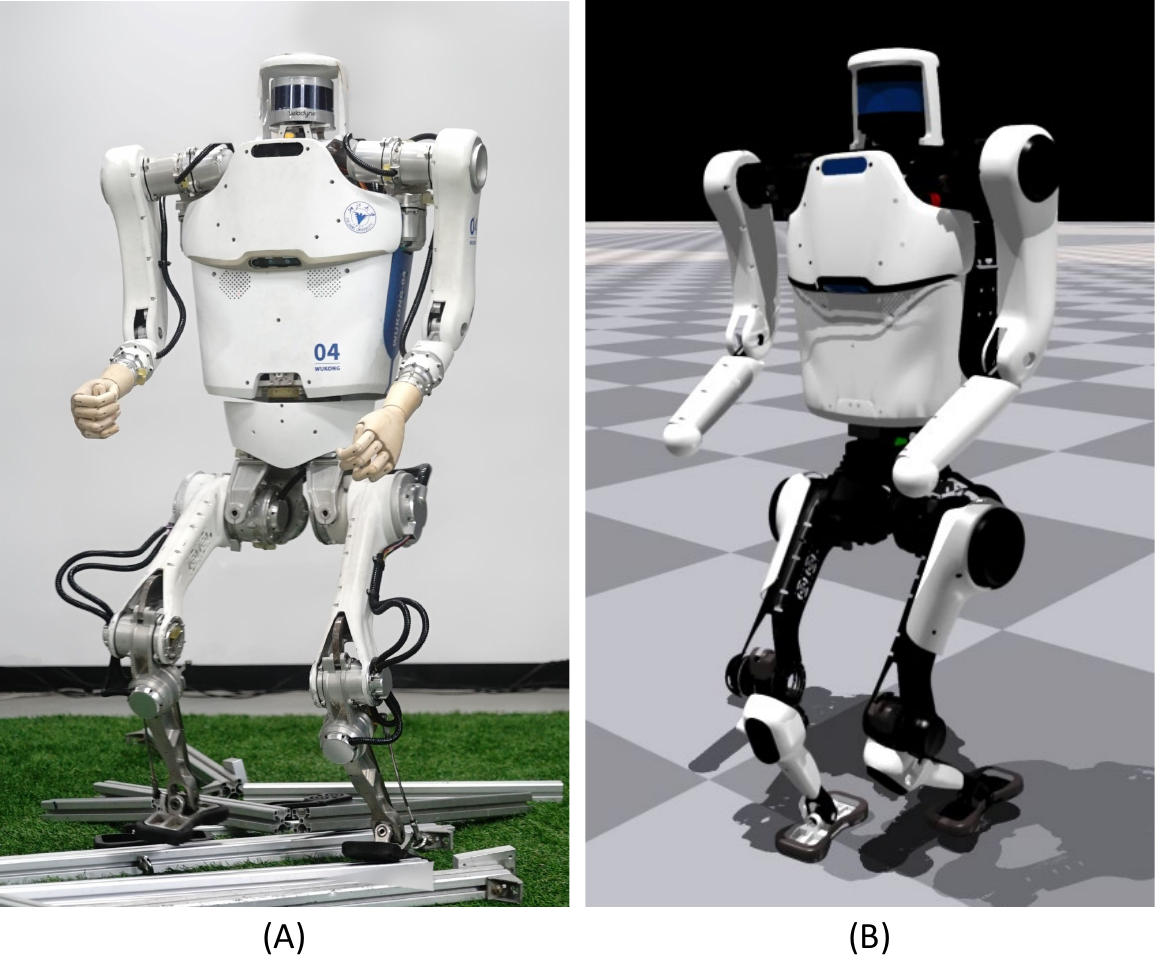}
    \caption{Wukong IV humanoid model. (A) Real-world Wukong-IV humanoid. (B) Simulated model in IsaacGym.}
    \label{Figure_Robot}
\end{figure}

The proposed training method is implemented on the Wukong-IV humanoid robot. It is 1.4 m tall, weighs 45 kg, and is actuated by 21 electric motor joints. The robot has 6 degrees of freedom (DoF) on each leg and 4 DoFs on each arm. A picture of a real Wukong-IV robot and its dynamics model in a simulation environment is shown in Fig.\ref{Figure_Robot}. The training environments are implemented in IsaacGym. The policies are built under PyTorch\cite{pytorch} framework. To prove that the proposed method is insensitive to random seeds, all mentioned policies are trained repeatedly for 6 times with CPU timestamp upon running as random seeds. 

\subsection{Saliency analysis}

Using saliency metrics, we systematically assess the impact of estimated values on the resultant actions. In our study, we begin by constructing a comprehensive estimation policy, a full estimation policy designed to encompass all potentially relevant states for humanoid locomotion tasks. Specifically, we incorporate explicit estimation terms pertaining to base linear velocity, base height, and the heightmap surrounding the feet, while incorporating fixed-width implicit information for subsequent saliency analysis. The saliency results are shown in Fig.\ref{Fig_saliency}.

According to the saliency experiments, we can see that the most crucial estimation is the linear velocity which takes 0.845 relative importance on average. This can be expected because the training prioritizes linear velocity velocity tracking. The second important estimation is heightmap around the feet, which is connected with the next motion decision tightly. Base height and implicit encoding showed a small influence on the final action.

\subsection{Comparison group setup}

After the saliency test, we have quantitatively understood the importance of different estimation terms to the action. In this section, we will further examine the influence on the final performance of the policies. The saliency tests showed that the velocity estimation has the biggest impact on the behavior of the policy, so we assume that estimating states with higher importance would improve the overall locomotion performance. To compare the performances of policies that adopt different estimations, we set up six experiment groups as follows:
\begin{itemize}

\item \textbf{EstimatorNet}(EstNet)\cite{estimatorNet}: This policy's encoder only works as an explicit estimator. It estimates body linear velocity without any implicit encoding or decoder to reconstruct the proprioceptive observation. 

\item \textbf{Velocity Estimation}(Key1): This policy estimates the body velocity along with a 16-dimension latent vector. The velocity proved to be the most important estimation in the saliency analysis, and we also left enough space for the policy itself to encode useful information from the proprioceptive history.

\item \textbf{Key Estimation}(Key2): This policy estimates the two most important states, base linear velocity, and heightmap around the feet, which are the top 2 key estimation states.

\item \textbf{Full Estimation}(FullEst): This policy's structure is the same as the policy used in saliency analysis. It estimates all mentioned states, including base linear velocity, heightmap surrounding the feet, and body height.

\item \textbf{Irrelevant Estimation}(IrrEst): This policy estimates the least important explicit state, the base height, along with a 16-dimension latent vector. 

\item \textbf{Implicit Encoding}(Implicit) This policy has no explicit estimation. The encoder gives only a 16-dimension latent vector. 

\end{itemize}

\begin{figure}
    \centering
    \includegraphics[width=8 cm]{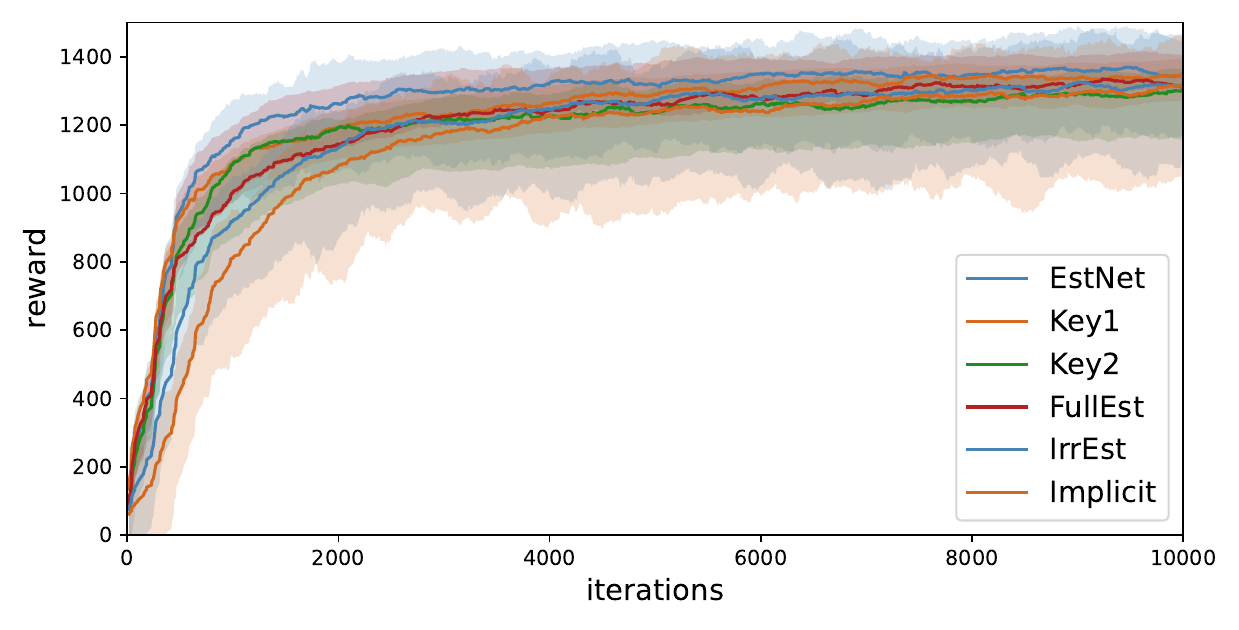}
    \caption{Reward plots of different trials. The solid lines are the filtered mean episode reward, and the shaded area marks the ranges of the episode reward values.}
    \label{Fig_reward}
\end{figure}

\begin{table*}[ht]
    \vspace{2mm}
    \centering
    \caption{Performance summary}
    \begin{center}
        
    \begin{tabular}{p{3cm}p{1.8cm}p{1.8cm}p{1.8cm}p{1.8cm}p{1.8cm}p{1.8cm}}
    \toprule
         
          & \textbf{EstNet} &\textbf{Key1} &\textbf{Key2} &\textbf{FullEst} &\textbf{IrrEst} &\textbf{Implicit}   \\
         \toprule
        $\Bar{r}_{final}$ & 1318 & \textbf{1347} & 1308 & 1324 & 1332 & 1313 \\
        $RMS(\Delta v_{sim})$  & 0.2247 & \textbf{0.2231} & 0.2330 & 0.2256 & 0.3892 & 0.4343 \\
        $RMS(\Delta g_{sim})$ & 0.0228 & 0.0184 & 0.0219 & 0.0187 & 0.0275 & \textbf{0.0162} \\
        $RMS(\Delta v_{real})$ & 0.2396 & 0.2671 & 0.2553 & \textbf{0.2270} & 0.3945 & 0.3919 \\
        $RMS(\Delta g_{real})$ & 0.0283 & 0.0180 & \textbf{0.0143} & 0.0281 & 0.0339 & 0.0258 \\
        ${TSR}_{stair}$ & 0.75 & 0.75 & \textbf{0.80} & 0.70 & 0.10 & 0.25 \\
        ${TSR}_{slope}$ & 0.85 & 0.90 & \textbf{0.95} & 0.90 & 0.80 & 0.75 \\
        ${TSR}_{metal}$ & 0.70 & \textbf{1.00} & \textbf{1.00} & \textbf{1.00} & 0.50 & 0.50 \\
        ${TSR}_{grass}$ & 0.95 & \textbf{1.00} & \textbf{1.00} & 0.95 & 0.80 & 0.90  \\
         \bottomrule
    \end{tabular}
    \end{center}
    \label{table_perf}
\end{table*}

\begin{figure}
    \centering
        \includegraphics[width=8cm]{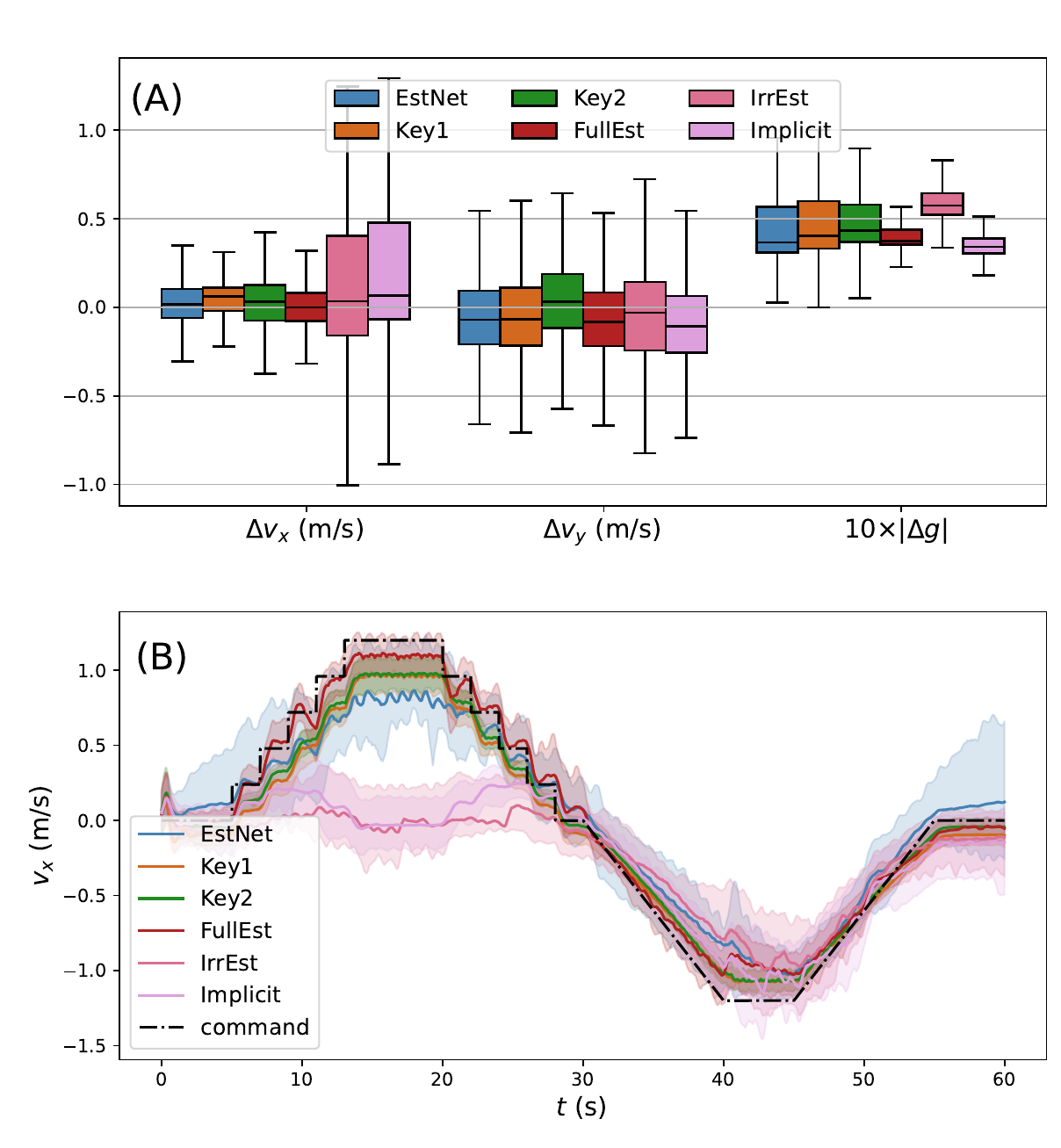}
    \caption{Velocity tracking plots of various estimation policies. (A) The figure presents a box plot illustrating the distribution of velocity and orientation tracking errors in a simulated environment. Orientation error is magnified tenfold for improved visibility. Each colored box denotes the interquartile range, spanning from the 25th to the 75th percentiles, with the horizontal line representing the median value. Additionally, error bars indicate the boundaries where $p < 0.05$. (B) This figure includes velocity plots showcasing the policies' real-world performance in tracking a predefined command trajectory (indicated by the black dashed plot). The shaded area encompasses the range of measured velocities across all repeated trials, while the solid lines represent the mean velocity values derived from all recorded trajectories.}
    \label{Fig_VelTrack}
\end{figure}

\begin{figure*}[ht]
    \centering
    \includegraphics[width=16cm]{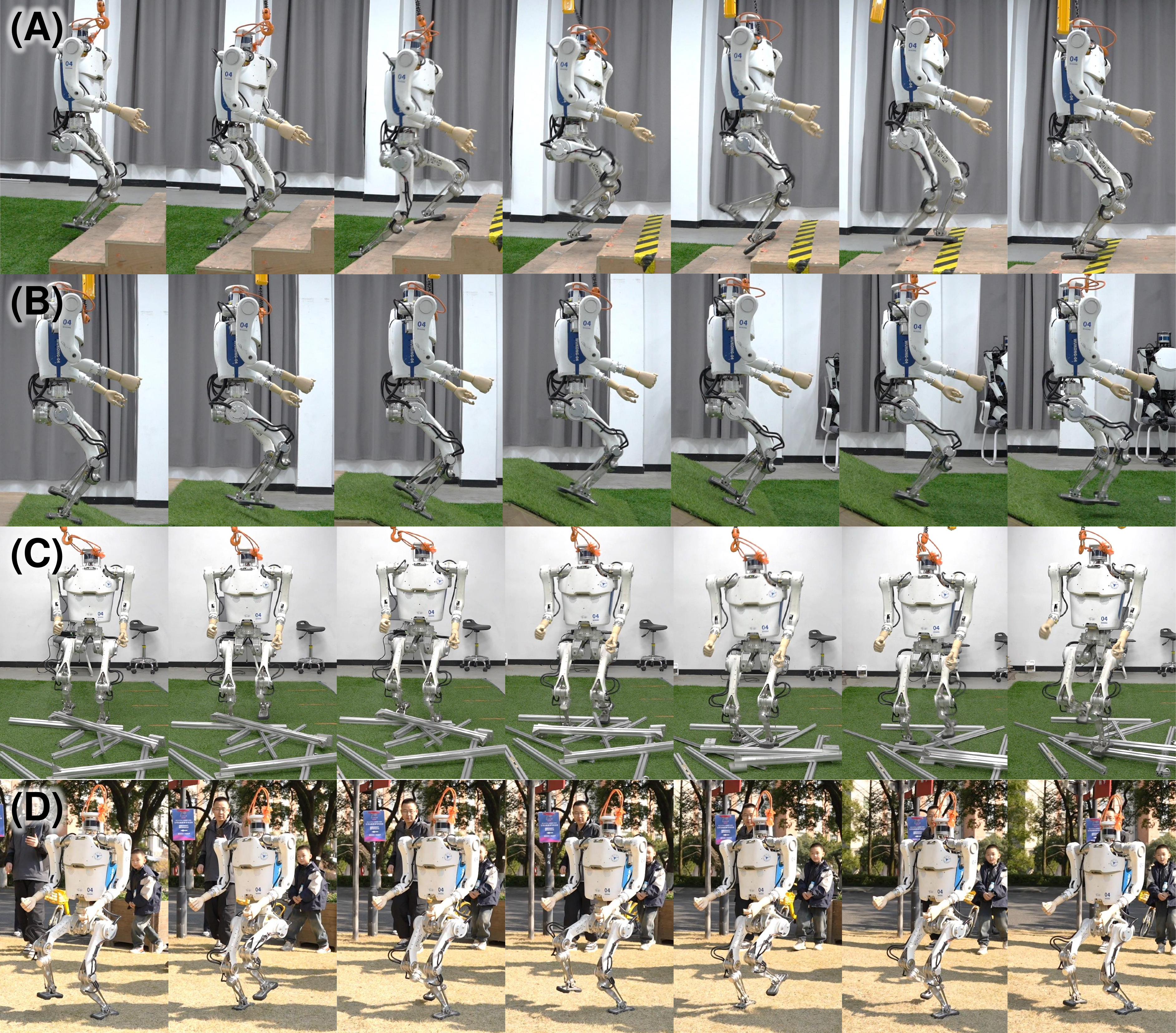}
    \caption{Snapshots of Wukong-IV humanoid traversing different terrains. (A) 3-step 10cm stairs. (B) $25\deg$ slope. (C) Indoor terrain with aluminum profiles. (D) Natural grass field. Check the supplementary video for more experiments.}
    \label{Fig_Snapshots}
\end{figure*}
\subsection{Performance metrics}

In evaluating the performance of trained policies for humanoid locomotion, we employ a comprehensive set of metrics to assess various aspects of task execution. 

we gauge the effectiveness of the learned policies by considering the final reward level at convergence, which is calculated as the mean episodic reward over the last 10 episodes of training. This metric provides insight into the overall success achieved by the policies in accomplishing the designated locomotion tasks. The velocity tracking error is denoted as $\Bar{r}_{final}$.

Furthermore, we scrutinize the policies' velocity tracking accuracy by comparing the root mean square (RMS) velocity error. In simulation, the error is computed by testing the policies with 1024 10-second random constant command trajectories on flat terrain. In the real world, the error is computed by testing the policies with a predefined 60-sec template velocity command trajectory with ramps and slopes 16 times in indoor settings, and in this case, LiDAR odometry serves as the measurement of body velocity. In both scenarios, the maximal velocity command is 1.2m/s. The velocity tracking error is denoted as $RMS(\Delta v_{sim/real})$.

Additionally, we evaluate the policies' orientation stability by quantifying the RMS body orientation fluctuation when tracking the aforementioned template velocity commands on flat ground. We define orientation fluctuation by the distance between the gravity vector and its nominal value. The orientation fluctuation metrics are denoted as $RMS(\Delta g_{sim/real})$.

Moreover, to ascertain the traversability and robustness of the policies, we measure the successful traversal rate over a variety of challenging terrains in the real world, including 3-step 10cm stairs, 25-degree slopes, terrain featuring random 4cm metal profiles, and natural grass fields. The successful rates are computed based on 20 trials on each type of terrain with a human operator sending velocity commands. These metrics collectively provide a comprehensive evaluation framework, allowing us to assess the policies' performance across diverse locomotion scenarios and environments. The successful rates are denoted as ${TSR}_{stair}, {TSR}_{slope},{TSR}_{metal}, {TSR}_{grass}$.

\subsection{Experiments and Analysis}

The reward curves of the training trials are shown in Fig.\ref{Fig_reward}. The plots depicted in the figure exhibit a convergence towards a reward level approximately averaging 1300. This observation suggests that all examined policies successfully acquire and refine fundamental locomotion skills. Although discernible variances in the average final reward values among different policy groups exist, these differences, with disparities of up to 2.98\%, become negligible when considering the influence of domain randomization on the mean episodic reward value. Consequently, these discrepancies are deemed inconsequential, and all policies are regarded as having undergone equivalent training. Thus, our comparative group design effectively safeguards the capacity of all policies to acquire and master locomotion skills.

Base velocity tracking performance is visualized in Fig.\ref{Fig_VelTrack}. A quantitative summary can also be found in Table.\ref{table_perf}. From Fig.\ref{Fig_VelTrack}(A) we can find out that the policy groups' sagittal velocity tracking performances largely vary, while the coronal performances resemble each other. This makes sense because the humanoid robot's feasible task space is too narrow for the robot to counteract the coronal velocity error and other disturbances. In the sagittal direction, the policies with irrelevant estimations and only implicit encodings have poor tracking performance compared with the other groups with explicit base velocity estimations. Also, we can note that the velocity error boxes are asymmetric. They have larger areas above zero, which means that the policies are worse at tracking large forward velocity commands compared with backward commands. The velocity tracking error distribution reveals that explicit velocity estimation plays a crucial role in learning humanoid locomotion tasks, and it is hard for the policies to learn to embed similar information into implicit vectors without direct supervision present.

The right column of Fig. \ref{Fig_VelTrack}(A) illustrates the distribution of orientation error. It is notable that policies employing implicit encoding mechanisms demonstrate superior pose stability when compared to alternative policies. However, upon closer examination of their velocity-tracking performance, it becomes evident that this heightened stability arises from a propensity to minimize movement. Thus, policies characterized by optimal orientation stability are those embracing full estimation strategies, which explicitly encompass a wider array of state variables.

Figure \ref{Fig_VelTrack}(B) portrays the velocity profiles of the comparison groups. Throughout this experiment, all policies track a pre-defined velocity trajectory composed of discrete steps and ramps. Consistent with previous observations regarding velocity-excluded policies, these two categories demonstrate a consistent inability to accurately track forward velocity commands, albeit exhibiting compliance with backward directives. Conversely, policies featuring explicit velocity estimation capabilities generally succeed in maintaining a velocity of 1.2m/s. Moreover, in addition to velocity tracking performance, this figure also sheds light on sim-to-real transfer proficiency. Notably, while EstimatorNet policies exhibit commendable tracking accuracy in simulation, a notable decline in accuracy is observed in real-world settings. Furthermore, plots associated with EstimatorNet policies possess a broader coverage area, indicative of heightened sim-to-real brittleness compared to counterparts employing implicit encoding strategies.

Regarding robustness, corresponding experimental findings are detailed in Table \ref{table_perf}. Policies characterized by Key2 estimations, representing the two most crucial estimation variables, demonstrate the highest levels of robustness across all four terrain types. These policies exhibit a remarkable ability to traverse complex environments, including stone curbs and 20cm platforms. The FullEst policies tend to overreact to unexpected obstacles and thus demonstrate inferior robustness than Key2 policies. Snapshots depicting the robot's traversal of various terrains can be found in Fig. \ref{Fig_Snapshots}, while additional experiment videos are available in the supplementary materials. Generally, Key2 policies perform the best in terms of tracking accuracy and real-world robustness.

\section{CONCLUSION}
\label{Conclusion}
In this work, we quantitatively inspected the importance of learned explicit estimations and evaluated the locomotion performance of different estimation designs. Among all of the estimated states, velocity emerges as the paramount factor, with heightmap ranking second. Policies equipped with velocity estimation exhibit enhanced locomotion capabilities, particularly evident in the precision of velocity tracking and the attainment of maximum speeds in both forward directions. Incorporating heightmap estimation bolsters adaptability to complex terrains, albeit with a lesser impact compared to sole velocity estimation. Regarding physical transferability, implicit encoding encompasses information not covered by explicit estimation, thereby enhancing policy adaptability during transitions from simulation to real-world environments.

There are some limits to be further studied. We have not considered the potentially complex effects of upper body movements on robots. If arm movements are substantial or involve carrying loads, inertial information might also influence performance. This inertial information may also require estimation. Besides, this policy is blind and doesn't include perception information. This work primarily focuses on the software and hardware configuration of the Wukong 4 robot, we acknowledge that additional information may indeed influence the distribution of the importance of state estimation.

\section*{ACKNOWLEDGMENT}

We are grateful to have Shixin Luo, Songbo Li, Shuaichen Zhang, and Guanxun Lang from the Robotics and Machine Intelligence Lab, Zhejiang University to help with real robot experiments. We also thank Zhiyong Tang, Kai Xu, and Xueyin Zhang from DeepRobotics Inc. for maintaining and repairing the robot hardware. Last but not least, we would like to express our gratitude to Alex Zhibin Li from University College London for the inspiration and suggestions about real robot deployment.


\section*{APPENDIX}

\begin{table}[!ht]
    \caption{Randomization Parameters}
    \centering
    \begin{tabular}{l c c c c}
    \hline
    \textbf{Parameter} & \textbf{Range} & \textbf{Unit} \\
    \hline
    Base CoM position & [-0.15,0.15] & m\\
    Base load mass & [-2.0, 12.5] & kg\\
    Friction rate & [0.25,1.25] & -\\
    Motor strength  & [0.8,1.2] & -\\
    Kp factor  & [0.9,1.1] & -\\
    Kd factor  & [0.9,1.1] & -\\
    latency & [0,2] & network step \\
    \hline
    \end{tabular}
    \label{Appendix_rand}
\end{table}

\begin{table}[!ht]
    \vspace{2mm}
    \centering
    \caption{Hyperparameters for PPO and neural network}
    \begin{tabular}{l c}
    \hline
    \textbf{Parameter} & \textbf{Value}\\
    \hline
    Number of environments & 4096 \\
    Learning epochs & 4 \\
    Initial learning rate & 5e-4 \\
    Gamma & 0.996 \\
    Lamda & 0.95 \\
    Number of batches & 4 \\
    Backbone hidden layers & [2048, 512, 128] \\
    Encoder hidden layers & [1024, 256, 64] \\
    Activation function & ELU \\
    Velocity loss coefficient & 1 \\
    Heightmap loss coefficient & 0.5 \\
    Body height loss coefficient & 2 \\
    VAE $\beta$ & 50 \\
    Prediction loss coefficient & 2 \\
    \hline
    \end{tabular}
    \label{Appendix_nn}
\end{table}

\vspace{12pt}

\bibliography{conference_101719}
\bibliographystyle{IEEEtran}

\end{document}